\documentclass[11pt]{article}

\usepackage{macros}
\usepackage{a4wide}
\usepackage{colacl}
\usepackage{multicol}  % llistes en multicolumnes
\usepackage{amssymb}   % math 
\usepackage{isolatin1} % accents

\include{epsf}

\author{J. Daud\'e, L. Padr\'o \& G. Rigau\\
        TALP Research Center\\
        Departament de Llenguatges i Sistemes Inform\`atics\\
        Universitat Polit\`ecnica de Catalunya. Barcelona\\
        {\tt \{daude,padro,g.rigau\}@lsi.upc.es} }
\title{Mapping Multilingual Hierarchies Using Relaxation Labeling}

\begin{document}
\maketitle

\begin{abstract}
This paper explores the automatic construction of a multilingual
Lexical Knowledge Base from pre-existing lexical resources. We present a new 
and robust approach for linking already existing lexical/semantic
hierarchies. We used a constraint satisfaction algorithm (relaxation
labeling) to select --among
all the candidate translations proposed by a bilingual dictionary--
the right English WordNet synset for each sense in a taxonomy 
automatically derived from a Spanish monolingual dictionary. Although
on average, there are 15 possible WordNet connections for each sense
in the taxonomy, the method achieves an accuracy over 80\%. Finally, 
we also propose several ways in which this technique
could be applied to enrich and improve existing lexical databases.
\end{abstract}

\section{Introduction}

There is an increasing need of having available general, accurate and
broad coverage multilingual lexical/semantic resources for developing 
 {\sc nl} applications. Thus, 
a very active field inside {\sc nl} during the last years has been the 
fast development of generic language resources.

Several attempts have been performed to produce multilingual ontologies. 
In \cite{ageno94}, a Spanish/English bilingual dictionary is used to
(semi)automatically link Spanish and English taxonomies extracted 
from {\sc dgile} \cite{dgile87} and {\sc ldoce} \cite{ldoce87}.
Similarly, a simple automatic approach for linking 
Spanish taxonomies extracted from {\sc dgile} to WordNet \cite{miller91} 
synsets is proposed in \cite{rigau95}. 
The work reported in \cite{knight94} focuses on the construction of Sensus, a 
large knowledge base for supporting the Pangloss machine translation system. 
In \cite{okumura94} (semi)automatic 
methods for associating a Japanese lexicon to an English ontology using a bilingual 
dictionary are described. Several experiments aligning {\sc edr} and 
WordNet ontologies are described in \cite{utiyama97}. Several lexical resources and 
techniques are combined in \cite{atserias97} to map Spanish words from a 
bilingual dictionary to WordNet, and in \cite{farreres98} the use of the 
taxonomic structure derived from a monolingual {\sc mrd} is proposed as an aid 
to this mapping process.

This paper presents a novel approach for merging already existing hierarchies. 
The method has been applied to attach substantial fragments of the Spanish 
taxonomy derived from {\sc dgile} \cite{rigau98} to the English WordNet 
using a bilingual dictionary for connecting both hierarchies. 

 This paper is organized as follows: In section \ref{application} we describe the 
used technique (the relaxation labeling algorithm) and its application to hierarchy
mapping. In section \ref{constraints} we describe the constraints used in the relaxation
process, and finally, after presenting some experiments and preliminary results, we 
offer some conclusions and outline further lines of research.

%%%%%%%%%%%%%%%%%%%%%%%%%%%%%%%%%%%%%%%%%%%%%%%%%%%%%%%%%
\section{Application of Relaxation Labeling to NLP}
\label{application}

Relaxation labeling (RL) is a generic name for a family of iterative
algorithms which perform function optimization, based on local
information.  See \cite{Torras89} for a summary.
Its most remarkable feature is that it can deal with any kind of constraints,
thus, the model can be improved by adding any constraints available, and
the algorithm is independent of the complexity of the model. That is,
we can use more sophisticated constraints without changing the
algorithm. 

The algorithm has been applied to {\sc pos} tagging 
\cite{Marquez97a}, shallow parsing \cite{Voutilainen97} and
to word sense disambiguation \cite{Padro98a}.

Although other function optimization algorithms 
could have been used (e.g. genetic algorithms, 
simmulated annealing, etc.), we found {\sc RL} to be 
suitable to our purposes, given its ability to use models
based on context constraints, and the existence of previous
work on applying it to {\sc nlp} tasks.

Detailed explanation of the algorithm can be found in \cite{Torras89},
while its application to NLP tasks, advantages and drawbacks are
addressed in \cite{Padro98a}.

%+++++++++++++++++++++++++++++++++++++++++++++++++++++%
\subsection{Algorithm Description}

The Relaxation Labeling algorithm deals with a set of variables (which may
represent words, synsets, etc.), each of which may take one among several 
different labels ({\sc pos} tags, senses, {\sc mrd} entries, etc.). 
There is also 
a set of constraints which state compatibility or incompatibility of a 
combination of pairs variable--label.

 The aim of the algorithm is to find a weight assignment for each
possible label for each variable, such that (a) the weights for 
the labels of the same variable add up to one, and (b) the weight
assignation satisfies --to the maximum possible extent-- the 
set of constraints. 

 Summarizing, the algorithm performs constraint satisfaction to solve
a consistent labeling problem. The followed steps are:

\begin{enumerate}
\item Start with a random weight assignment.
\item Compute the {\sl support} value for each label of each variable. 
       Support is computed according to the constraint set and to the
       current weights for labels belonging to context variables.
\item Increase the weights of the labels more compatible with the
      context (larger support) and decrease those of
      the less compatible labels (smaller support). Weights are changed
      proportionally to the support received from the context. 
\item If a stopping/convergence criterion is satisfied,
      stop, otherwise go to step 2. We use the
      criterion of stopping when there are no more changes, although
      more sophisticated heuristic procedures may also be used to stop
      relaxation processes \cite{Eklundh78,Richards81}. 
\end{enumerate}

The cost of the algorithm is proportional to the product of the number of 
variables by the number of constraints.

%+++++++++++++++++++++++++++++++++++++++++++++++++++++%
\subsection{Application to taxonomy mapping}

   As described in previous sections, the problem we are dealing
  with is to map two taxonomies.
  That is:
    \begin{itemize}
       \item The starting point is a sense disambiguated Spanish taxonomy 
             --automatically extracted from a monolingual dictionary 
             \cite{rigau98}--.
       \item We have a conceptual taxonomy (e.g. WordNet \cite{miller91}), in which the nodes
             represent concepts, organized as {\sl synsets}.
       \item We want to relate both taxonomies in order to have an
             assignation of each sense of the Spanish taxonomy to a WN synset.
    \end{itemize} 

  The modeling of the problem is the following:
   \begin{itemize}
      \item Each sense in the Spanish taxonomy is a variable for the relaxation
           algorithm.
      \item The possible labels for that variable, are all the WN synsets which
           contain a word that is a possible translation of the Spanish sense.
           Thus, we will need a bilingual dictionary to know all the possible  
           translations for a given Spanish word. This has the effect
           of losing the sense information we had in the Spanish taxonomy.
      \item The algorithm will need constraints stating whether a synset is 
           a suitable assignment for a sense. These constraints will
           rely on the taxonomy structure. Details are given in 
           section~\ref{constraints}.
   \end{itemize}

%%%%%%%%%%%%%%%%%%%%%%%%%%%%%%%%%%%%%%%%%%%%%%%%%%%%%%%%%
\section{The Constraints}
\label{constraints}

   Constraints are used by relaxation labeling algorithm to increase or decrease the
weight for a variable label. In our case, constraints increase the
weights for the connections between a sense in the Spanish taxonomy and a
WordNet synset. Increasing the weight for a connection implies decreasing
the weights for all the other possible connections for the same node.
 To increase the weight for a connection, constraints look for already connected 
nodes that have the same relationships in both taxonomies.

   Although there is a wide range of relationships between WordNet synsets which 
can be used to build constraints, we have focused on the hyper/hyponym relationships.
That is, we increase the weight for a connection when the involved nodes have
hypernyms/hyponyms also connected. We consider hyper/hyponym relationships 
either directly or indirectly (i.e. ancestors or descendants), depending on
the kind of constraint used.

   Figure~\ref{f-exemple1} shows an example of possible connections between two
taxonomies. Connection $C_4$ will have its weight increased due to $C_5$, $C_6$ and
$C_1$, while connections $C_2$ and $C_3$ will have their weights decreased.

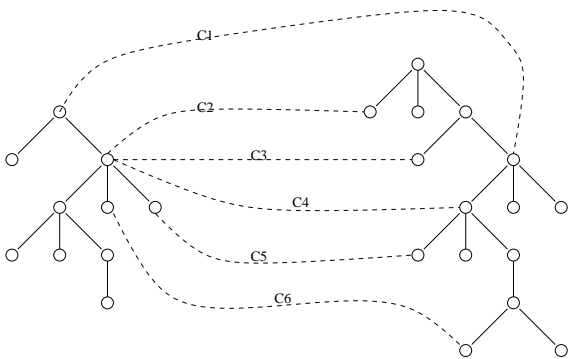
\figeps{exemple1}{Example of connections between taxonomies.}{f-exemple1}

  Constraints are coded with three characters {\sc xyz}, which are
read as follows: The last character, {\sc z}, indicates whether the 
constraints requires the existence of a connected hypernym ({\sc e}), 
hyponym ({\sc o}), or both ({\sc b}). The two first characters indicate
how the hyper/hyponym relationship is considered in the Spanish taxonomy (character
{\sc x}) and in WordNet (character {\sc y}): ({\sc i}) indicates that only 
{\sl immediate} 
hyper/hyponym match, and ({\sc a}) indicates that {\sl any} ancestor/descendant matches.

  Thus, we have constraints {\sc iie}/{\sc iio} which increase the
weight for a connection between a Spanish sense and a WordNet synset when there
is a connection between their respective hypernyms/hyponyms. Constraint {\sc iib}
requires the {\sl simultaneous} satisfaction of {\sc iie} and {\sc iio}.

  Similarly, we have constraints {\sc iae}/{\sc iao}, which 
increase the weight for a connection between a Spanish sense and a WordNet 
synset when there is a connection between the immediate hypernym/hyponym of
the Spanish sense and any ancestor/descendant of the {\sc wn} synset. Constraint
{\sc iab} requires the {\sl simultaneous} satisfaction of {\sc iae} and {\sc iao}.
Symmetrically, constraints {\sc aie}, {\sc aio} and {\sc aib}, admit recursion
on the Spanish taxonomy, but not in WordNet.

  Finally, constraints {\sc aae}, {\sc aao} and {\sc aab}, admit recursion on both sides.

For instance, the following example shows a taxonomy in which the {\sc iie}
constraint would be enough to connect the Spanish node {\sl rapaz} 
to the {\sl\lt bird\_of\_prey\gt} synset, given that there is a 
connection between {\sl ave} (hypernym of {\sl rapaz}) and 
{\tt animal} {\sl\lt bird\gt} (hypernym of {\sl\lt bird\_of\_prey\gt}).
%%\vspace{-0.3cm}
\begin{tabbing} 
xxxxxx \= \kill
animal \> \fletxa ({\tt Tops} {\sl\lt animal,animate\_being,...\gt})\\
       \> \fletxa ({\tt person} {\sl\lt beast,brute,...\gt})        \\
       \> \fletxa ({\tt person} {\sl\lt dunce,blockhead,...\gt})    \\
xxxx \= \kill
     \> ave \= \fletxa ({\tt animal} {\sl\lt bird\gt})               \\
     \>     \> \fletxa ({\tt animal} {\sl\lt fowl,poultry,...\gt})   \\
     \>     \> \fletxa ({\tt artifact} {\sl\lt bird,shuttle,...\gt}) \\
     \>     \> \fletxa ({\tt food} {\sl\lt fowl,poultry,...\gt})     \\
     \>     \> \fletxa ({\tt person} {\sl\lt dame,doll,...\gt})      \\
xxxxxx \= \kill
         \> faisan \= \fletxa ({\tt animal} {\sl\lt pheasant\gt})   \\
         \>        \> \fletxa ({\tt food} {\sl\lt pheasant\gt})     \\
         \> rapaz \= \fletxa ({\tt animal} {\sl\lt bird\_of\_prey,...\gt}) \\
         \>       \> \fletxa ({\tt person} {\sl\lt cub,lad,...\gt})               \\
         \>       \> \fletxa ({\tt person} {\sl\lt chap,fellow,...\gt})           \\
         \>       \> \fletxa ({\tt person} {\sl\lt lass,young\_girl,...\gt})      \\
\end{tabbing}
%%\vspace{-0.4cm}
Constraint {\sc iie} would --wrongly-- connect the Spanish sense 
{\sl faisán} to the {\tt food} {\sl\lt pheasant\gt} synset, since 
there is a connection between its immediate hypernym ({\sl ave}) and 
the immediate hypernym {\tt food} {\sl\lt pheasant\gt} (which 
is {\tt food} {\sl\lt fowl,poultry,...\gt}), but the {\tt animal} synsets
for {\sl ave} are non--immediate ancestors of the {\tt animal} synsets
for {\sl\lt pheasant\gt}. This would be rightly solved when using 
{\sc iae} or {\sc aae} constraints.

  More information on constraints and their application can be found in
\cite{Daude99a}.

%%%%%%%%%%%%%%%%%%%%%%%%%%%%%%%%%%%%%%%%%%%%%%%%%%%%%%%%%
\section{Experiments and Results}
\label{experiments}

  In this section we will describe a set of experiments and the results obtained. 
A brief description of the used resources is included to set the reader in the 
test environment.

%+++++++++++++++++++++++++++++++++++++++++++++++++++++%
\subsection{Spanish Taxonomies}

 We tested the relaxation labeling algorithm with the described constraints on
a set of disambiguated Spanish taxonomies automatically acquired from
monolingual dictionaries. These taxonomies were automatically assigned
to a WordNet semantic file \cite{rigau97,rigau98}.  We tested
the performance of the method on two different kinds of taxonomies:
those assigned to a well defined and concrete semantic files 
({\tt noun.animal}, {\tt noun.food}), and those assigned to more abstract and
less structured ones ({\tt noun.cognition} and {\tt noun.communication}). 

 We performed experiments directly on the taxonomies extracted by \cite{rigau97}, as well
as on slight variations of them. Namely, we tested on the following {\sl modified} 
taxonomies:
\begin{description}
\item[+top] Add a new virtual top as an hypernym of all the top nodes of taxonomies
      belonging to the same semantic file. The virtual top is connected to the
      top synset of the WordNet semantic file. In this way, all the
      taxonomies assigned to a semantic file, are converted to a single one. 
\item[no-senses] The original taxonomies were built taking into account dictionary entries.
       Thus, the nodes are not words, but dictionary {\sl senses}. This test consists of
       collapsing together all the sibling nodes that have the same word, regardless of
       the dictionary sense they came from. This is done as an attempt to minimize 
       the noise introduced at the sense level by the taxonomy building procedure.
\end{description}

%\begin{table*}[htb] \centering
%\begin{tabular}{l|cccc|}
%   & {\tt noun.animal} & {\tt noun.food} & {\tt noun.cognition} & {\tt noun.communication} \\ \hline
%original &   1675      &    746          &   1494               &     2241           \\
%+top     &   1676      &    747          &   1495               &     2242           \\
%no-senses&   1600      &    696          &   1402               &     2063           \\ \hline
%\end{tabular}
%\caption{Number of nodes in each test taxonomy.}
%\label{t-nnodes}
%\end{table*}

%+++++++++++++++++++++++++++++++++++++++++++++++++++++%
\subsection{Bilingual dictionaries}

   The possible connections between a node in the Spanish taxonomy and WN 
synsets were extracted from bilingual dictionaries. Each node has as candidate 
connections all the synsets for all the words that are possible translations 
for the Spanish word, according to the bilingual dictionary.
Although the Spanish taxonomy nodes are dictionary senses, bilingual
dictionaries translate words. Thus, this step introduces noise 
in the form of irrelevant connections,  since not all translations 
necessarily hold for a single dictionary sense.

  We used an integration of several bilingual sources available. This
multi--source dictionary contains 124,949 translations (between 53,830
English and 41,273 Spanish nouns).

 Since not all words in the taxonomy appear in our bilingual dictionaries, 
coverage will be partial.  Table \ref{t-traduc} shows the percentage of nodes in each 
taxonomy that appear in the dictionaries (and thus, that may be 
connected to WN). 

\begin{table*}[htb] \centering
\begin{tabular}{l|ccc|}
                          &  original & +top  & no-senses  \\  \hline
{\tt noun.animal}         &   45\%  & 45\%   &  43\%  \\   
{\tt noun.food}           &   55\%  & 56\%   &  52\%  \\   
{\tt noun.cognition}      &   54\%  & 55\%   &  52\%  \\   
{\tt noun.communication}  &   66\%  & 66\%   &  64\%  \\ \hline 
\end{tabular}
\caption{Percentage of nodes with bilingual connection in each test taxonomy.}
\label{t-traduc}
\end{table*}

   Among the words that appear in the bilingual dictionary, some have 
 only one candidate connection --i.e. are {\sl monosemous}--.
 Since selecting a connection for these cases is trivial, we will focus on the 
 {\sl polysemic} nodes. Table \ref{t-ambigues} 
 shows the percentage of polysemic nodes (over the number of 
 words with bilingual connection) in each test taxonomy. The average 
 polysemy ratio (number of candidate connections per Spanish sense) is
 15.8, ranging from 9.7 for taxonomies in {\tt noun.animal},
 to 20.1 for less structured domains such as {\tt noun.communication}.

\begin{table*}[htb] \centering
\begin{tabular}{l|ccc|}
                          & original & +top  & no-senses  \\  \hline
{\tt noun.animal}         &  77\%  & 77\%   &  75\%  \\   
{\tt noun.food}           &  81\%  & 81\%   &  79\%  \\   
{\tt noun.cognition}      &  74\%  & 74\%   &  72\%  \\   
{\tt noun.communication}  &  87\%  & 87\%   &  86\%  \\ \hline 
\end{tabular}
\caption{Percentage of nodes with more than one candidate connection.}
\label{t-ambigues}
\end{table*}

%+++++++++++++++++++++++++++++++++++++++++++++++++++++%
\subsection{Results}

  In the performed tests we used simultaneously all 
constraints with the same recursion pattern. This yields the packs: {\sc ii*},
{\sc ai*}, {\sc ia*} and {\sc aa*}, which were applied to all the 
taxonomies for the four test semantic files.

  Table \ref{t-res-coverage} presents coverage figures for the different test sets,
computed as the amount of nodes for which some constraint is applied and thus their weight
assignment is changed. Percentage is given over the total amount of nodes with bilingual
connections.

\begin{table*}[htb] \centering
\begin{tabular}{l|l|rrrr|}
  WN file & taxonomy & \multicolumn{1}{c}{{\sc ii*}} &  \multicolumn{1}{c}{{\sc ai*}}  &  \multicolumn{1}{c}{{\sc ia*}}  &  \multicolumn{1}{c|}{{\sc aa*}}  \\ \hline
                  & original & 134 (23\%) &  135 (23\%) &  357 (62\%) & 365 (63\%) \\
{\tt noun.animal} & +top     & 138 (24\%) &  143 (25\%) &  375 (65\%) & 454 (78\%) \\
                  & no-senses& 118 (23\%) &  119 (20\%) &  311 (61\%) & 319 (62\%) \\ \hline
                 & original & 119 (36\%) &  130 (39\%) & 164 (49\%) & 180 (63\%) \\
{\tt noun.food}  & +top     & 134 (40\%) &  158 (47\%) & 194 (58\%) & 259 (77\%) \\
                 & no-senses& 102 (36\%) &  111 (39\%) & 153 (51\%) & 156 (55\%) \\ \hline
                     & original & 225 (37\%) &  230 (38\%) &  360 (60\%) & 373 (62\%) \\
{\tt noun.cognition} & +top    & 230 (38\%) &  240 (40\%) &  395 (65\%) & 509 (84\%) \\
                     & no-senses& 192 (37\%) &  197 (38\%) &  306 (59\%) & 318 (61\%) \\ \hline
                         &original & 552 (43\%) &  577 (45\%) &  737 (57\%) & 760 (59\%) \\
{\tt noun.communication} &+top     & 589 (46\%) &  697 (54\%) &  802 (62\%) &1136 (88\%) \\
                         &no-senses& 485 (43\%) &  509 (45\%) &  645 (57\%) & 668 (59\%) \\ \hline
\end{tabular}
\caption{Coverage of each constraint set for different test sets.}
\label{t-res-coverage}
\end{table*}

  To evaluate the precision of the algorithm, we hand checked the 
results for the {\sl original} taxonomies, using {\sc aa*} constraints.
Precision results can be divided in several cases, depending on the
correctness of the Spanish taxonomies used as a starting point.
\begin{description}
\item[$T_{OK}, F_{OK}$] The Spanish taxonomy was well built and correctly assigned to the semantic file.
\item[$T_{OK}, F_{NOK}$] The Spanish taxonomy was well built, but wrongly assigned to the semantic file.
\item[$T_{NOK}$] The Spanish taxonomy was wrongly built. 
\end{description}

  In each case, the algorithm selects a connection for each sense, we will count 
how many connections are right/wrong in the first and second cases. In the third
case the taxonomy was wrongly extracted and is nonsense, so the assignations
cannot be evaluated.

  Note that we can distinguish right/wrong assignations in the second case 
because the connections are taken 
into account over the whole WN, not only on the semantic file being processed.
 So, the algorithm may end up correctly assigning the words of a hierarchy, 
even when it was assigned to the wrong semantic file. For instance, in the hierarchy
%%\vspace{-0.2cm}
\begin{tabbing} 
xx \= xxxx \= xxxx \= \kill 
     \> piel ({\sl skin, fur, peel, pelt})\\
     \>     \> \fletxa  marta ({\sl sable, marten, coal\_back}) \\
     \>     \> \fletxa  vison ({\sl mink, mink\_coat})\\
\end{tabbing}
%%\vspace{-0.4cm}
\noindent all words may belong either to the semantic file {\tt noun.substance} (senses 
related to {\sl fur, pelt}) or to {\tt noun.animal} ({\sl animal, animal\_part} senses),
among others. The right {\tt noun.substance} synsets for each word are selected, since
there was no synset for {\sl piel} that was ancestor of
the {\tt animal} senses of {\sl marta} and {\sl visón}.

 In this case, the hierarchy was well built, and well solved by the algorithm. The 
only mistake was having assigned it to the {\tt noun.animal} semantic file, so we will
count it as a right choice of the relaxation labeling algorithm, but write it
in a separate column.

  Tables \ref{t-res-precisio1} and \ref{t-res-precisio2} show the precision 
rates for each {\sl original} taxonomy. 
 In the former, figures are given over polysemic 
words (nodes with more than one candidate connection). In the later, 
figures are computed overall (nodes with at least
one candidate connection). 

Accuracy is computed at the semantic file level, i.e., if a word is 
assigned a synset of the right semantic file, it is computed as right,
otherwise, as wrong. 

To give an idea of the task complexity and the quality of the reported results, 
even with this simplified evaluation, consider the following: 

\begin{itemize}
\item Those nodes with only one possible synset for the right 
      semantic file (30\% in average, ranging from 22\% in 
      {\tt noun.communication} to 45\% in {\tt noun.animal}) are not affected
      by the evaluation at the semantic file level.
\item The remaining nodes have more than one possible synset in the right 
      semantic file: 6.3 in average (ranging from 3.0 for {\tt noun.animal} 
      to 8.7 for {\tt noun.communication}).
\item Thus, we can consider that we are evaluating a task easier than the actual one
      (the actual evaluation would be performed at the synset level). This 
      simplified task has an average polysemy of 6.7 possible choices per sense, while
      the actual task at the synset level would have 15.8. Although this situates
      the baseline of a random assignment about 15\% instead of 6\%, it
      is still a hard task.
\end{itemize}

\begin{table*}[htb] \centering
\begin{tabular}{l|ccc|c|}
                   & precision over    & precision over    & total precision & number       \\
                   & $T_{OK}, F_{OK}$  & $T_{OK}, F_{NOK}$ & over $T_{OK}$   & of $T_{NOK}$ \\ \hline
{\tt animal}       & 279 (90\%) &  30  (91\%) & 309 (90\%) &  23  \\
{\tt food}         & 166 (94\%) &   3 (100\%) & 169 (94\%) &   2  \\
{\tt cognition}    & 198 (67\%) &  27  (90\%) & 225 (69\%) &  49  \\
{\tt communication}& 533 (77\%) &  40  (97\%) & 573 (78\%) &  16  \\ \hline
\end{tabular}
\caption{Precision results over polysemic words for the test taxonomies.}
\label{t-res-precisio1}
\end{table*}

\begin{table*}[htb] \centering
\begin{tabular}{l|ccc|}
                   & precision over    & precision over    & total precision  \\
                   & $T_{OK}, F_{OK}$  & $T_{OK}, F_{NOK}$ & over $T_{OK}$   \\ \hline
{\tt animal}       & 424 (93\%) &  62  (95\%) & 486 (93\%)      \\
{\tt food}         & 166 (94\%) &  83 (100\%) & 149 (96\%)      \\
{\tt cognition}    & 200 (67\%) & 245  (99\%) & 445 (82\%)      \\
{\tt communication}& 536 (77\%) & 234  (99\%) & 760 (81\%)      \\ \hline
\end{tabular}
\caption{Precision results over all words for the test taxonomies.}
\label{t-res-precisio2}
\end{table*}

%%%%%%%%%%%%%%%%%%%%%%%%%%%%%%%%%%%%%%%%%%%%%%%%%%%%%%%%%
\section{Conclusions}

   We have applied the relaxation labeling algorithm to assign an appropriate
WN synset to each node of an automatically extracted taxonomy. Results
for two different kinds of conceptual structures have been reported,
and they point that this may be an accurate and robust method (not
based on ad-hoc heuristics) to connect hierarchies (even in different 
languages).

   The experiments performed up to now seem to indicate that:
\begin{itemize}
  \item The relaxation labeling algorithm is a good technique to link
     two different hierarchies. For each node with several possible connections,
     the candidate that best matches the surrounding structure is selected.
  \item The only information used by the algorithm are the
     hyper/hyponymy relationships in both taxonomies. These local constraints
     are propagated throughout the hierarchies to produce a global solution.
  \item There is a certain amount of noise in the different phases of the
     process. First, the taxonomies were automatically acquired and assigned
     to semantic files. Second, the bilingual dictionary translates words, not
     senses, which introduces irrelevant candidate connections.
  \item The size and coverage of the bilingual dictionaries used to establish
     the candidate connections is an important issue. A dictionary with larger 
     coverage increases the amount of nodes with candidate connections and thus the
     algorithm coverage
\end{itemize}

%%%%%%%%%%%%%%%%%%%%%%%%%%%%%%%%%%%%%%%%%%%%%%%%%%%%%%%%%
\section{Proposals for Further Work}

   Some issues to be addressed to improve the algorithm performance are the following:
\begin{itemize}
  \item Further test and evaluate the precision of the algorithm. In this 
      direction we plan --apart from performing wider hand checking of the 
      results, both to file and synset level-- to use the presented technique to 
      link WN1.5 with WN1.6. Since there is already a mapping between both 
      versions, the experiment would provide an idea of the accuracy of the  
      technique and of its applicability to different hierarchies of
      the same language. 
      In addition, it would constitute an easy way to update existing lexical resources.
  \item Use other relationships apart from hyper/hyponymy to build constraints
      to select the best connection (e.g. sibling, cousin, synonymy, meronymy, etc.).
  \item To palliate the low coverage of the bilingual dictionaries, candidate
      translations could be inferred from connections of surrounding senses. For
      instance, if a sense has no candidate connections, but its hypernym does, 
      we could consider as candidate connections for that node all the hyponyms 
      of the synset connected to its hypernym.
  \item Use the algorithm to enrich the Spanish part of EuroWordNet taxonomy.
      It could also be applied to include taxonomies for other languages 
      not currently in the {\sc ewn} project.
\end{itemize}

  In addition, some ideas to further exploit the possibilities of these techniques are:
\begin{itemize}
  \item Use {\sc ewn} instead of {\sc wn} as the target taxonomy. This would
      largely increase the coverage, since the candidate connections missing
      in the bilingual dictionaries could be obtained from the Spanish part 
      of {\sc ewn}, and viceversa. In addition, it
      would be useful to detect {\sl gaps} in the Spanish part of {\sc ewn}, since 
      a {\sc ewn} synset with no Spanish words in {\sc ewn}, could be assigned one
      via the connections obtained from the bilingual dictionaries. 
  \item Since we are connecting dictionary senses (the entries in 
    the {\sc mrd} used to build the taxonomies) to {\sc ewn} synsets:
    First of all, we could use this to disambiguate the right sense for the genus 
      of an entry. For instance, in the Spanish taxonomies, the genus for the
      entry {\sl queso}\_1 (cheese) is {\sl masa} (mass) but this word has several
      dictionary entries. Connecting the taxonomy to {\sc ewn}, we would be able
      to find out which is the appropriate sense for {\sl masa}, and thus, which
      is the right genus sense for {\sl queso}\_1.
       Secondly, once we had each dictionary sense connected to a {\sc ewn} synset,
      we could enrich {\sc ewn} with the definitions in the {\sc mrd}, using them
      as Spanish glosses.
    \item Map the Spanish part of {\sc ewn} to {\sc wn1.6}. This could be 
       done either directly, or via mapping {\sc wn1.5--wn1.6}.
\end{itemize}

\section{Acknowledgments}

This research has been partially funded by the Spanish Research Department (ITEM Project TIC96-1243-C03-03), the Catalan Research Department (CREL project), and the UE Commission (EuroWordNet LE4003).

\bibliographystyle{acl}
\bibliography{/usr/usuaris/ia/padro/articles/fullbib}

\end{document}